# Convolutional Transformer Neural Collaborative Filtering

Pang Li [1], Shahrul Azman Mohd Noah[2], Hafiz Mohd Sarim[3]
Center for Artificial Intelligence and Technology
Faculty of Information Science and Technology
The National University of Malaysia
a184459@siswa.ukm.edu.my[1], shahrul@ukm.edu.my[2], hms@ukm.edu.my[3]

**ABSTRACT**

In this study, we introduce Convolutional Transformer Neural Collaborative Filtering (CTNCF), a novel approach aimed at enhancing recommendation systems by effectively capturing high-order structural information in user-item interactions. CTNCF represents a significant advancement over the traditional Neural Collaborative Filtering (NCF) model by seamlessly integrating Convolutional Neural Networks (CNNs) and Transformer layers. This sophisticated integration enables the model to adeptly capture and understand complex interaction patterns inherent in recommendation systems. Specifically, CNNs are employed to extract local features from user and item embeddings, allowing the model to capture intricate spatial dependencies within the data. Furthermore, the utilization of Transformer layers enables the model to capture long-range dependencies and interactions among user and item features, thereby enhancing its ability to understand the underlying relationships in the data. To validate the effectiveness of our proposed CTNCF framework, we conduct extensive experiments on two real-world datasets. The results demonstrate that CTNCF significantly outperforms state-of-the-art approaches, highlighting its efficacy in improving recommendation system performance.

## 1. INTRODUCTION

In the development of personalized recommendation systems, creating personalized predictive models has been widely adopted to help individuals identify content of interest. These systems collect various features of individuals, such as demographic characteristics, ratings of items (explicit feedback), or interactions between users and items (implicit feedback) (Aljunid, M. F., & Huchaiah, M. D. 2022). Their goal is to provide future preferences based on past interactions, and they have been widely used in various fields such as e-commerce and online streaming services. The most direct technique for generating recommendations is Collaborative Filtering (CF) (Li et al. 2019).

Although CF has demonstrated its effectiveness, it often falls short in capturing the complex interactions between users and items. With the emergence of Neural Collaborative Filtering (NCF) (He et al. 2017), a new possibility for recommendation systems has been provided by modeling the non-linear relationships between users and items using deep learning. However, despite progress in understanding complex interactions, there is still room for improvement in capturing high-order structural information present in observed data (Wu, L., He, X., Wang, X., Zhang, K., & Wang, M. 2022), which is crucial for further enhancing recommendation accuracy and user satisfaction.

In this work, we propose a novel approach called Convolutional Transformer Neural Collaborative Filtering (CTNCF) aimed at improving recommendation systems by capturing high-order structural information in user-item interactions. Our method enhances the traditional NCF model by integrating Convolutional Neural Networks (CNN) and Transformer layers. This integration enables the model to adeptly capture complex interaction patterns. Specifically, CNN is used to extract local features from user and item embeddings, allowing the model to capture spatial dependencies. Additionally, Transformer layers are employed to capture long-term dependencies and interactions between user and item features. Extensive experiments on two real-world datasets demonstrate that our proposed CTNCF framework significantly outperforms NCF method.

The main contributions of this work are:

1. Introduction of a new recommendation system approach—CTNCF, which enhances the capability of traditional NCF models by integrating CNN and Transformer layers to capture complex interaction patterns between users and items.
2. Utilization of CNN to capture local features and Transformer layers to capture long-term dependencies, aiding in understanding the complex relationships between user behavior and item features.
3. Experiments conducted on two real-world datasets demonstrate that CTNCF has significant advantages in improving recommendation quality compared to existing state-of-the-art models.

The structure of this paper is as follows: The first section provides an introduction to recommendation systems. The second section discusses related work. The third section presents the proposed CTNCF method in detail. The fourth section covers the preparation of experiments, including datasets and evaluation methods. The fifth section includes experimental results and discussions. Finally, the sixth section summarizes the work and discusses future directions.

## 2. RELATED WORKS

In this section, we recall previous research related to collaborative filtering and generative models.

### 2.1 Collaborative Filtering

Collaborative filtering (CF) stands as a cornerstone method for developing recommendation systems. It posits that a user can be effectively represented by aggregating their interactions with various items, and vice versa. The fundamental element in implementing collaborative filtering is the user-item interaction matrix. Initial approaches, as referenced in studies (Shi et al. 2014), rely on matrix factorization (MF) to concurrently model the latent dimensions of both users and items. With advancements in deep learning and graph neural networks (GNN), CF methods have evolved significantly. Wu et al. (2022) explored optimizing user-item interactions through deep learning, enhancing recommendation precision with autoencoders. Additionally, Chen et al. (2023) proposed an attention-based model that more effectively captures the dynamics and complexity of user behavior. This method employs multi-head attention networks to analyze nonlinear relationships between user history and item attributes, thus improving recommendation quality. Regarding GNN applications, Chen et al. (2022) introduced a novel GNN framework designed to handle sparse user-item interaction data. This framework improves information transmission efficiency through advanced aggregation and update rules, allowing the model to perform effectively even with limited interaction data. Moreover, Liu et al. (2022) developed a graph-based CF method that incorporates edge reconstruction techniques to optimize latent connections between users and items, enhancing the recommendation system's interpretability and accuracy. With the success of contrastive learning in computer vision and natural language processing, this technique has also been incorporated into CF. Zhang et al. (2024) demonstrated a CF framework that integrates contrastive learning, effectively enhancing the model's generalization ability and robustness by learning from positive and negative user-item pairs.

Moreover, in terms of content-enhanced CF models, Yu et al. (2024) proposed a method that integrates textual and visual features for recommendations. By deeply analyzing user reviews and product images, this model provides more personalized and accurate recommendations. Additionally, Khan, W. et al. (2023) utilized cutting-edge natural language processing technologies, like BERT and GPT, to analyze the semantics of user reviews, refining detailed user preferences and item characteristics.

**2.2 Generative Models**

Generative models have increasingly become a vital component of modern recommender systems. Initial research in this area primarily utilized Variational Autoencoders (VAEs) (Marcuzzo et al. 2022) and Generative Adversarial Networks (GANs) (Deldjoo et al. 2021) to enhance recommendations by generating new user-item interaction data. More recent developments have incorporated diffusion models (Yang et al. 2023), which offer novel ways to generate item recommendations by simulating the process of gradually denoising data. Furthermore, advancements in large language models (LLMs) have paved the way for new approaches in recommendations (Yang et al. 2023). Several studies have experimented with the capabilities of LLMs by employing techniques like parameter-efficient tuning or instruction-based tuning to tailor recommendations. Some researchers have also transformed various recommendation scenarios into unified tasks of natural language generation, optimizing these models through multi-task learning frameworks such as P6 (Ngo et al. 2020). A notable development is the TIGER method (Rajput et al. 2024), which utilizes an RQ-VAE for constructing generative identification descriptors, followed by employing encoder-decoder based transformers for sequential recommendation. Yuan et al. (2023) explored building generative identification descriptors from a pretrained SASRec model for similar purposes.

Despite these innovations, generative recommendation systems still face challenges. A primary concern is the ineffective integration of collaborative filtering signals with item content information. The quest to effectively combine these elements within a generative sequence-to-sequence framework remains an open and significant research challenge in the field.

## 3. METHODOLOGY

We first analyze the drawbacks of the multilayer perceptron (MLP) network used for the baseline model NCF. To overcome these shortcomings, we propose the Convolutional Transformer Neural Collaborative Filtering (CTNCF) model. Subsequently, we exhaustively elaborate the technical details employed in each layer of the CTNCF model.

### 3.1 Drawbacks of MLP

In recommender systems, MLP (He et al. 2017) is particularly prone to overfitting problems when dealing with sparse and high-dimensional data, and although it performs well on training data, its performance on unknown data is unsatisfactory. In addition, MLP architectures demand a large number of parameters as the network deepens or widens, which not only increases computational cost and memory requirements, but also limits their scalability on large datasets (Alzubaidi et al. 2021). In terms of feature interaction modeling, although MLP is capable of modeling nonlinear relationships between features, it underperforms in capturing more complex hierarchies or interactions between different types of data, which is particularly important in collaborative filtering, where interactions between users and commodities are critical. Generalization capability is also another challenge for MLP, especially when dealing with new users or new goods (i.e., cold-start problems), where less information may be difficult to convey effectively due to the reliance on intensive representation learning (Cong et al. 2023). These drawbacks limit the potential of MLP to be applied in modern recommender systems.

### 3.2 The CTNCF Method

To address these shortcomings of MLP, we propose a novel architecture, the Convolutional Transformer Neural Collaborative Filtering (CTNCF) model, which integrates Convolutional Neural Networks (CNNs) and a transformer layer. This hybrid architecture utilizes CNNs to reduce the model's sensitivity to overfitting through local sensing fields and weight sharing, which enhances the model's generalization ability on out-of-sample data. Meanwhile, the transformer layer flexibly handles various data dependencies through its attention mechanism, which enriches the model's ability to understand complex user-item interactions in large-scale

datasets.The CTNCF model is designed to optimize both local and global feature extraction, which not only solves the challenges of modeling hierarchical and spatial dependencies, but also improves the processing power for large-scale data processing. This enhancement is achieved without a substantial increase in computational overhead, as the model's data processing efficiency can be augmented through the utilization of contemporary hardware architectures and parallel computing techniques. In addition, the integration of CNN and transformer techniques helps to better adapt to new users or items, potentially alleviating the cold-start problem and thus providing a powerful solution to the limitations of traditional MLP-based collaborative filtering systems.

### 3.3 CTNCF Framework

Figure 1 illustrates the CTNCF framework. The target of modeling is to estimate the matching score between user $u$ and item $i$, i.e., $\hat{y}_{ui}$; and then we can generate a personalized recommendation list of items for a user based on the scores.

**Input and Embedding Layer.** For each user $u$ and item $i$, along with their properties such as IDs, the model first applies one-hot encoding. The feature vectors $v_U^u$ and $v_I^i$ for user $u$ and item $i$, respectively, are transformed into dense representations through embedding matrices:

$$p_u = P^T v_U^u, \quad q_i = Q^T v_I^i, \tag{1}$$

where P and Q represent the embedding matrices for users and items, $p_u$ and $q_i$ are the embedding vectors for the user and item, respectively. The symbol T stands for transpose. Transposing a matrix or vector means flipping its dimensions. If you transpose a column vector, it becomes a row vector, and vice versa.

**Concatenation Layer.** This layer is pivotal in synthesizing the features extracted from the embeddings. It utilizes a blend of matrix factorization (MF) for capturing linear interaction patterns and convolutional neural networks (CNN) for identifying nonlinear, intricate details within the data. Matrix factorization effectively pinpoints the fundamental user-item affinities, while CNNs delve deeper to uncover complex, localized interactions. The outputs from these methods are then concatenated to form an enriched interaction profile that merges both linear and nonlinear insights.

**Matrix Factorization (MF) Component.** The matrix factorization interaction is modeled using a simple but effective mathematical operation that represents the dot product of user and item embeddings. This operation generates an initial interaction matrix via

$$E_{MF} = p_u \otimes q_i = p_u q_i^T, \qquad (2)$$

the resulting matrix $E_{MF}$ serves as the foundational structure for further enhancement. This matrix captures the primary latent interactions between user and item features but is limited to linear relationships. To enhance the model's capability, further complex non-linear transformations are applied.

**Convolutional Neural Network (CNN) Component.** This involves applying a series of convolutional filters to the embeddings, which helps in extracting localized feature interactions:

$$CNN_{user} = Conv1D(p_u), \; CNN_{item} = Conv1D(q_i), \qquad (3)$$

the convolution operations are designed to capture more complex and subtle patterns within the data, which are not typically captured by matrix factorization alone.

**Pooling Component.** Following convolution, a max pooling operation is applied to the outputs of the CNN layers. This reduces the dimensionality of the feature maps while retaining the most significant features. Pooling operations are essential for summarizing the feature information, making the subsequent layers less computationally intensive while focusing on the most informative parts of the interaction data.

**Transformer with Attention Layer.** To enhance the interaction map's capability to capture deeper relational patterns, the transformer architecture with self-attention is applied:

$$Attention(Q,K,V) = softmax(QK^T/\sqrt{d_k})V, \qquad (4)$$

where Q, K, V are queries, keys, and values derived from user and item embeddings, and $d_k$ is the scaling factor. This layer effectively models complex dependencies and interactions, enriching the representation before final prediction.

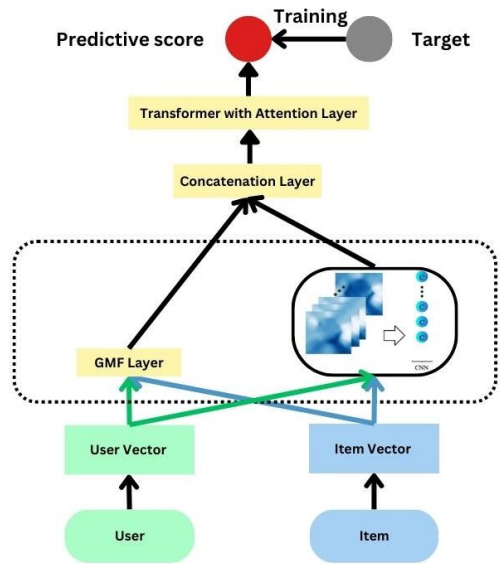

Figure 1: Convolutional Transformer Neural Collaborative Filtering model

## 4. Experiment

To completely evaluate our proposed methodology, we conducted a series of experiments to explore the following research questions:

**RQ1** Do our proposed CTNCF outperform the state-of-the-art methods?

**RQ2** How do the key hyperparameter in CNN (i.e., number of filters) affect CTNCF's performance?

**RQ3** Are the proposed Transfer with Attention layer helpful for learning from user-item interaction data and improving the recommendation performance?

### 4.1 Data Descriptions.

We conduct experiments using two prominent datasets in the domain of recommendation systems: MovieLens-1M and Amazon Electronics (AEle).

**MovieLens-1M**. The MovieLens-1M dataset consists of 1,000,209 movie ratings from 6,040 users on 3,706 movies. Each rating ranges from 1 to 5, where a higher value indicates greater user preference. Alongside ratings, the dataset provides additional information such as movie genres and user demographics. Due to its moderate size and density, MovieLens-1M is widely adopted as a benchmark dataset for evaluating recommendation algorithms.

**Amazon Electronics (AEle)**. The AEle dataset encompasses 7,824,482 ratings assigned by users to various electronics products available on the Amazon platform. Similar to MovieLens-1M, ratings in AEle are graded on a scale from 1 to 5. Notably, AEle is distinguished by its larger size and higher sparsity compared to MovieLens-1M, rendering it suitable for assessing algorithms' performance on large and sparsely populated datasets.

Table 1: Statistics of the evaluation datasets.

| Dataset | #Users | #Items | #Ratings | #Density(%) |
|---|---|---|---|---|
| MovieLens 1M | 6,040 | 3,706 | 1,000,209 | 4.4685 |
| Amazon(Electro) | 1,540 | 48,190 | 125,871 | 0.1700 |

## 4.2 Evaluation Protocols

We employ cross-validation to evaluate the performance of the recommendation system. To accomplish this, we randomly partition each user's historical interactions into training, validation, and test sets, with a ratio of 7:2:1. To gauge the efficacy of the model, we employ two prevalent metrics: recall@n and normalized discounted cumulative gain (NDCG@n). Recall is quantified by the formula:

$$\text{Recall}@n = \frac{\text{Number of relevant items in top n recommendations}}{\text{Total number of relevant items}}, \quad (5)$$

this measures the proportion of ground-truth items that appear in the top-n recommendation list. NDCG is calculated using the formul:

$$\text{NDCG}@n = \frac{DCG@n}{IDCG@n}, \quad (6)$$

where DCG@n (Discounted Cumulative Gain) accounts for the position of the relevant items by assigning higher scores to hits at higher ranks, and IDCG@n (Ideal Discounted Cumulative Gain) is the maximum possible DCG@n, representing the perfect ranking scenario. NDCG evaluates both the presence and the rank order of the recommended items. Each experiment is executed thrice, ensuring the robustness of results, and the mean score is reported, capturing the model's overall performance.

## 4.3 Baselines

To justify the effectiveness of our proposed CTNCF, we study the performance of the following methods:

1. **MLP** (He et al., 2017). This traditional neural network model is widely used in recommendation systems to model the user-item interaction function. By employing multiple layers of neurons, the MLP can learn complex non-linear relationships in data. It excels in scenarios where the interactions are not easily linearized, capturing subtleties in large and diverse datasets through its deep network structure.
2. **NCF** (He et al., 2017). This enhances the traditional matrix factorization (MF) model by integrating it with a multi-layer perceptron to learn the non-linear interactions between users and items. It combines the advantages of MF's latent factor approach for capturing the global structure of the data with deep neural networks' ability to model non-linearities, resulting in improved recommendation performance, especially in capturing complex and nuanced user preferences.
3. **LightFM**. This is a hybrid model that combines both collaborative filtering (CF) and content-based (CB) approaches. LightFM incorporates user and item metadata within the matrix factorization framework, allowing it to handle both explicit and implicit feedback effectively. This model is particularly useful in scenarios where there is sparse data or cold-start problems, as it leverages additional information beyond user-item interactions.

    We selected these methods for comparison because they are classic methods widely used in the field of recommendation systems and share similar model structures and objectives with our proposed CTNCF. By comparing the performance of CTNCF with these methods, we can better evaluate its superiority in recommendation tasks.

## 4.4 Implementation Details

In our experiments, we meticulously tuned a series of hyperparameters for our CTNCF model to optimize performance. The model's learning rate was set to 0.001, based on empirical evidence and preliminary experiments, to ensure effective and stable convergence. For regularization coefficients, we chose 0 for the matrix factorization part, considering the complexity of other parts of the model, while we applied a regularization coefficient of 0.01 for the convolutional neural network part to prevent overfitting. The embedding dimensions for the MF part was set to 4. Additionally, the configuration of the CNN layer includes 8, 16, 32, 64, and 128 filters and a kernel size of 3, which aids the model in learning local patterns of item and user characteristics. The introduction of Transformer layers further enhances the model's capabilities, with two Transformer layers set up to strengthen the model's understanding of user-item interaction data through attention mechanisms. The selection of these hyperparameters reflects our decision-making and trade-offs in model design, aiming to enhance the accuracy and relevance of the recommendation system by leveraging the advantages of MF, CNN, and Transformer layers collectively.

## 5. RESULTS AND DISCUSSION

In this section, we systematically analyze the performance of our proposed CTNCF model against several state-of-the-art methods. We also explore the influence of critical hyperparameters within the CTNCF framework and assess the efficacy of our novel Transformer with Attention layer.

### 5.1 Performance Comparison (RQ1)

Table 2 presents the Top-k recommendation performance metrics for our Convolutional Transformer Neural Collaborative Filtering (CTNCF) model alongside established methods such as MLP, NCF, and LightFM on the MovieLens-1M and AEle datasets. Performance is evaluated based on Recall@k and NDCG@k where k is set to 5, 10, and 20. These metrics provide insight into the accuracy and ranking quality of the recommended items. The results clearly demonstrate that CTNCF outstrips the baseline methods across all evaluated metrics. On the MovieLens-1M dataset, CTNCF achieves a notable increase in Recall@k, going as high as 0.0695 for k=5, compared to the next best performer, LightFM, which achieves 0.0653. Similarly, for NDCG@k, CTNCF shows a top score of 0.4288 for k=20, indicating its superior ability to rank highly relevant items at the top of the recommendation list. The AEle dataset reflects a consistent trend with CTNCF surpassing other methods, particularly in the NDCG@k metric, where the model's scores are demonstrably higher, reaching up to 0.2937 for k=20. This is a robust indicator of CTNCF's ranking effectiveness. Moreover, the RI (Relative Improvement) column summarizes the relative improvements of CTNCF over the baseline methods. The improvement ranges from +1.8% to +6.5%, with CTNCF exhibiting a substantial enhancement over the traditional NCF model. These results empirically affirm the advantages of integrating convolutional and Transformer layers into the collaborative filtering process, enabling the CTNCF model to adeptly capture complex interaction patterns and dependencies within user-item interaction data.

Table 2: Top-k recommendation performance where k ∈ {5, 10, 20}
*RI indicates the relative improvement of CTNCF over the baseline.

| Dataset | Metric | MLP | NCF | LightFM | CTNCF | RI |
|---|---|---|---|---|---|---|
| MovieLens-1M | Recall@5 | 0.0614 | 0.0647 | 0.0653 | **0.0695** | +6.43% |
| | NDCG@5 | 0.3710 | 0.3872 | 0.3886 | **0.4121** | +6.05% |
| | Recall@10 | 0.1143 | 0.1162 | 0.1197 | **0.1261** | +5.35% |
| | NDCG@10 | 0.3795 | 0.3891 | 0.3905 | **0.4115** | +5.38% |
| | Recall@20 | 0.1917 | 0.1989 | 0.2014 | **0.2107** | +4.62% |
| | NDCG@20 | 0.3995 | 0.4016 | 0.4078 | **0.4288** | +5.14% |
| AEle | Recall@5 | 0.0551 | 0.0585 | 0.0594 | **0.0615** | +3.54% |
| | NDCG@5 | 0.2437 | 0.2488 | 0.2599 | **0.2712** | +4.35% |
| | Recall@10 | 0.0988 | 0.1053 | 0.1052 | **0.1085** | +3.14% |
| | NDCG@10 | 0.2583 | 0.2602 | 0.2692 | **0.2795** | +3.82% |
| | Recall@20 | 0.1813 | 0.1847 | 0.1938 | **0.1974** | +1.86% |
| | NDCG@20 | 0.2697 | 0.2710 | 0.2845 | **0.2937** | +3.23% |

## 5.2 Influence of CNN Hyperparameters (RQ2)

Figure 2 and Figure 3 depict the effects of varying the number of CNN filters on the performance of the CTNCF model, measured by Recall@10 and NDCG@10, for the MovieLens-1M and AEle datasets, respectively. In Figure 2, we observe that Recall@10 for the MovieLens-1M dataset shows a positive trend as the number of filters increases, peaking at 64 filters before plateauing. This suggests an optimal number of filters for capturing the relevant features in the data without overfitting. Conversely, for the AEle dataset, performance peaks at 32 filters and then sharply declines, indicating a possible overparameterization that may lead to model overfitting for this particular dataset. Similarly, Figure 3 shows the performance pattern for NDCG@10 across both datasets with varying numbers of filters. The MovieLens-1M dataset displays a steady increase in NDCG@10 with the number of filters, achieving the best performance at 64 filters, and then stabilizes. On the other hand, the AEle dataset demonstrates an initial increase up to 32 filters, followed by a decrease, reinforcing the earlier observation that a more complex model does not necessarily translate to better performance on all datasets.

These figures underscore the significance of appropriate hyperparameter tuning, where the optimal number of filters is clearly contingent upon the characteristics of the dataset being used. The contrasting trends between the two datasets highlight the nuanced role that model complexity plays in the recommendation system's performance and emphasize the necessity of dataset-specific considerations when configuring CNN layers within the CTNCF framework.

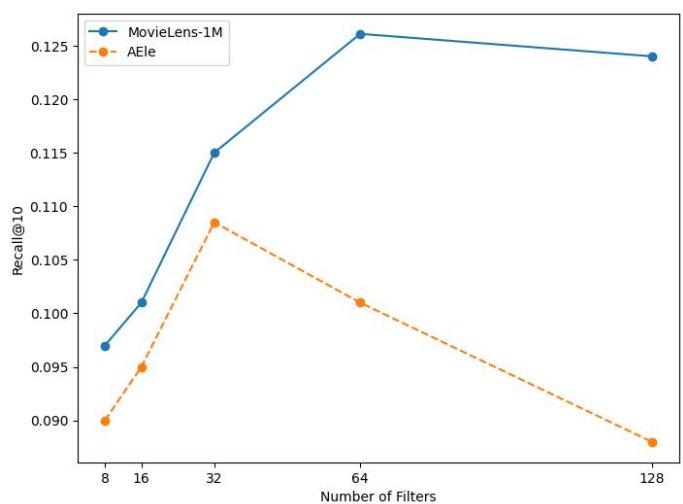

Figure 2: Variation in Recall@10 with CNN Filters for MovieLens-1M and AEle datasets

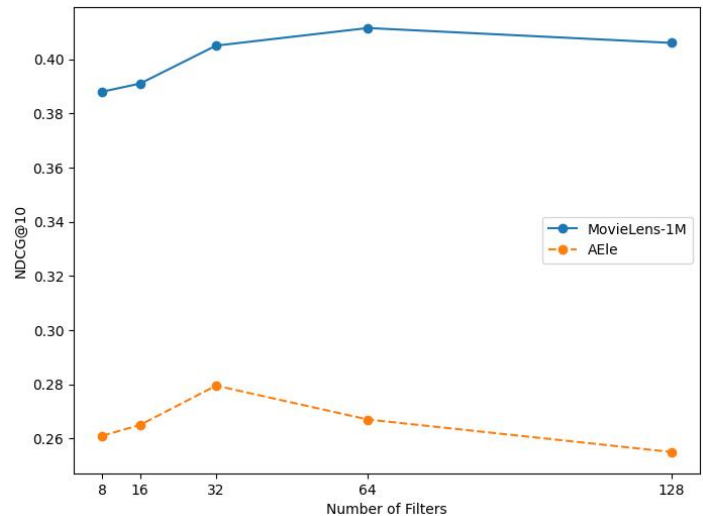

Figure 3: Variation in NDCG@10 with CNN Filters for MovieLens-1M and AEle datasets

**5.3 Effectiveness of the Transfer with Attention Layer (RQ3)**

Figure 4 and Figure 5 depict the enhancements in recommendation quality achieved by incorporating the Transformer with Attention Layer into the CTNCF model, evaluated using the Recall@10 and NDCG@10 metrics, respectively, on the MovieLens-1M and AEle datasets. The bar charts in Figure 4 reveal that the application of the Transformer with Attention Layer leads to a tangible improvement in Recall@10 for both datasets. Notably, the AEle dataset sees a substantial increase, suggesting the model's augmented capacity to unearth complex dependencies between user-item interactions. While the MovieLens-1M dataset also benefits from the attention layer, the gains are more modest, which may signal intrinsic data characteristics or an interaction pattern that is less influenced by the model's enhanced attention capabilities. Similarly, Figure 5 showcases the positive impact of the attention layer on the NDCG@10 metric. For the both datasets, there are a significant improvement in ranking accuracy, underscoring the attention layer's effectiveness in discerning the significance of different features for prediction. Collectively, these figures substantiate the utility of the Transformer with Attention Layer in the CTNCF model. They illustrate that, while there are dataset-specific responses to the incorporation of the attention mechanism, the overall trend points to enhanced performance, especially in discerning and prioritizing the nuances of user-item interactions for recommendations.

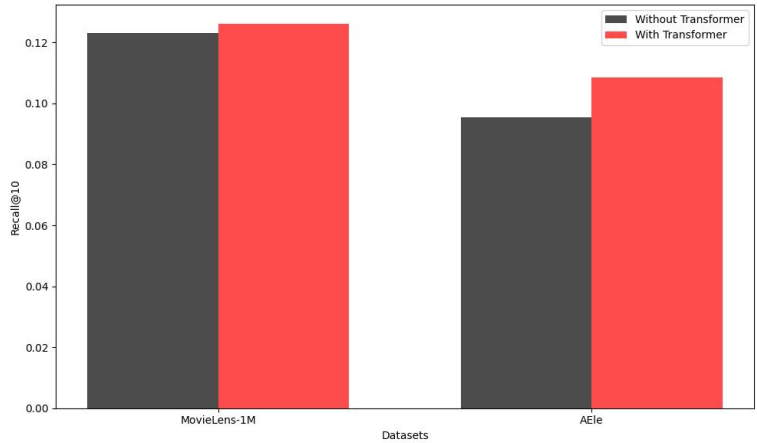

Figure 4: Impact of the Transformer with Attention Layer on Recall@10
for MovieLens-1M and AEle Datasets

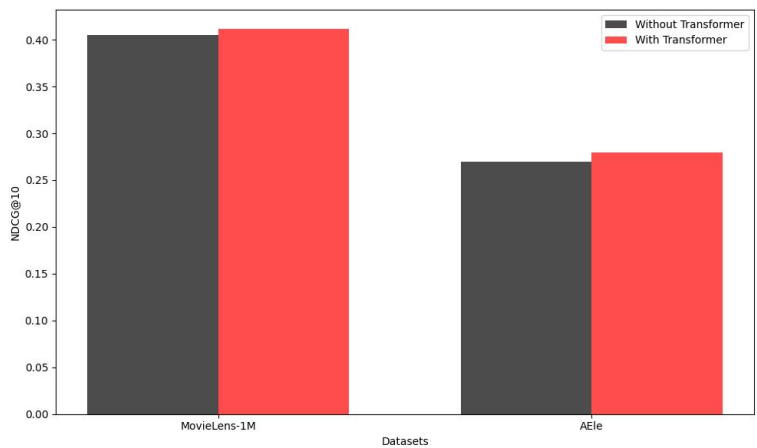

Figure 5: Impact of the Transformer with Attention Layer on NDCG@10
for MovieLens-1M and AEle Datasets

## 6. CONCLUSION AND FUTURE WORKS

In this work, we introduced a novel approach called Convolutional Transformer Neural Collaborative Filtering (CTNCF) to enhance recommendation systems by effectively capturing intricate structural information within user-item interactions. By integrating Convolutional Neural Networks (CNNs) and Transformer layers, our model, CTNCF, offers a sophisticated enhancement to the traditional Neural Collaborative Filtering (NCF) framework. Our approach leverages CNNs to extract local features from user and item embeddings, facilitating the capture of spatial dependencies within the data. Additionally, Transformer layers are employed to capture long-range dependencies and interactions among user and item features, enabling the model to discern complex interaction patterns more effectively. Extensive experiments conducted on real-world datasets demonstrated the superior performance of our CTNCF framework over state-of-the-art methods. In the future, we will explore the integration of ResNet and Transformer to optimize the collaborative filtering model, enhancing the model's feature extraction and global understanding capabilities. This strategy not only maintains the model's strong local analysis ability but also utilizes global information to further improve performance. For example, models like SpikingResformer (Shi et al. 2023) demonstrate the potential of this integration strategy in achieving high performance with low parameters. Additionally, we will also design a new loss function that explores both the informatand pair-wise loss as well as implicit and explicit feedback.

## ACKNOWLEDGEMENT

I am deeply grateful to my supervisors, Prof.Shahrul Azman Mohd Noah and Dr. Hafiz Mohd Sarim, for their invaluable guidance, advice, and support.

## REFERENCES

Aljunid, M. F., & Huchaiah, M. D. (2022). IntegrateCF: Integrating explicit and implicit feedback based on deep learning collaborative filtering algorithm. *Expert Systems with Applications, 207*, 117933.

Alzubaidi, L., Zhang, J., Humaidi, A. J., Al-Dujaili, A., Duan, Y., Al-Shamma, O., ... & Farhan, L. (2021). Review of deep learning: concepts, CNN architectures, challenges, applications, future directions. *Journal of big Data, 8*, 1-74.

Chen, L., Xie, T., Li, J., & Zheng, Z. (2022). Graph enhanced neural interaction model for recommendation. *Knowledge-Based Systems, 246*, 108616.

Cheng, H., Liu, M., Chen, L., Broszio, H., Sester, M., & Yang, M. Y. (2023). Gatraj: A graph-and attention-based multi-agent trajectory prediction model. *ISPRS Journal of Photogrammetry and Remote Sensing, 205*, 163-175.

Cong, S., & Zhou, Y. (2023). A review of convolutional neural network architectures and their optimizations. *Artificial Intelligence Review, 56*(3), 1905-1969.

Deldjoo, Y., Noia, T. D., & Merra, F. A. (2021). A survey on adversarial recommender systems: from attack/defense strategies to generative adversarial networks. *ACM Computing Surveys (CSUR), 54*(2), 1-38.

He, X., Liao, L., Zhang, H., Nie, L., Hu, X., & Chua, T. S. (2017, April). Neural collaborative filtering. *In Proceedings of the 26th international conference on world wide web* (pp. 173-182).

Khan, W., Daud, A., Khan, K., Muhammad, S., & Haq, R. (2023). Exploring the frontiers of deep learning and natural language processing: A comprehensive overview of key challenges and emerging trends. *Natural Language Processing Journal*, 100026.

Li, G., Yuchi, J., Yang, H., & Li, K. (2019). A network delay factor model based on the hidden Markov Model and Latent Dirichlet Allocation. *IEEE Access*, 7, 133136-133144.

Liu, M., Li, J., Liu, K., Wang, C., Peng, P., Li, G., ... & Xie, W. (2022). Graph-ICF: Item-based collaborative filtering based on graph neural network. *Knowledge-Based Systems, 251*, 109208.

Marcuzzo, M., Zangari, A., Albarelli, A., & Gasparetto, A. (2022). Recommendation systems: An insight into current development and future research challenges. *IEEE Access, 10*, 86578-86623.

Ngo, T., Kunkel, J., & Ziegler, J. (2020, July). Exploring mental models for transparent and controllable recommender systems: a qualitative study. *In Proceedings of the 28th ACM Conference on User Modeling, Adaptation and Personalization* (pp. 183-191).

Rajput, S., Mehta, N., Singh, A., Hulikal Keshavan, R., Vu, T., Heldt, L., ... & Sathiamoorthy, M. (2024). Recommender systems with generative retrieval. *Advances in Neural Information Processing Systems,* 36.

Shaheed, K., Qureshi, I., Abbas, F., Jabbar, S., Abbas, Q., Ahmad, H., & Sajid, M. Z. (2023). EfficientRMT-Net—An Efficient ResNet-50 and Vision Transformers Approach for Classifying Potato Plant Leaf Diseases. *Sensors, 23*(23), 9516.

Shi, X., Hao, Z., & Yu, Z. (2024). SpikingResformer: Bridging ResNet and Vision Transformer in Spiking Neural Networks. *In Proceedings of the IEEE/CVF Conference on Computer Vision and Pattern Recognition* (pp. 5610-5619).

Shi, Y., Larson, M., & Hanjalic, A. (2014). Collaborative filtering beyond the user-item matrix: A survey of the state of the art and future challenges. *ACM Computing Surveys (CSUR), 47*(1), 1-45.

Wu, L., He, X., Wang, X., Zhang, K., & Wang, M. (2022). A survey on accuracy-oriented neural recommendation: From collaborative filtering to information-rich recommendation. *IEEE Transactions on Knowledge and Data Engineering, 35*(5), 4425-4445.

Yang, J., Jin, H., Tang, R., Han, X., Feng, Q., Jiang, H., ... & Hu, X. (2023). Harnessing the power of llms in practice: A survey on chatgpt and beyond. ACM Transactions on Knowledge Discovery from Data.

Yang, L., Zhang, Z., Song, Y., Hong, S., Xu, R., Zhao, Y., ... & Yang, M. H. (2023). Diffusion models: A comprehensive survey of methods and applications. *ACM Computing Surveys, 56*(4), 1-39.

Yu, J., Tao, D., Wang, M., & Rui, Y. (2014). Learning to rank using user clicks and visual features for image retrieval. *IEEE transactions on cybernetics, 45*(4), 767-779.

Yuan, Z., Yuan, F., Song, Y., Li, Y., Fu, J., Yang, F., ... & Ni, Y. (2023, July). Modality-based recommender models revisited. *In Proceedings of the 46th International ACM SIGIR Conference on Research and Development in Information Retrieval* (pp. 2639-2649).

Zhang, A., Sheng, L., Cai, Z., Wang, X., & Chua, T. S. (2024). Empowering Collaborative Filtering with Principled Adversarial Contrastive Loss. *Advances in Neural Information Processing Systems, 36*.